\documentclass{article}

     \usepackage[preprint]{neurips_2020}

\usepackage[utf8]{inputenc} 
\usepackage[T1]{fontenc}    
\usepackage{hyperref}       
\usepackage{url}            
\usepackage{booktabs}       
\usepackage{amsfonts}       
\usepackage{nicefrac}       
\usepackage{microtype}      

\usepackage{amssymb}
\usepackage{algorithm}
\usepackage{algorithmic}
\usepackage{amsmath}
\usepackage{subfigure}

\usepackage{multirow}
\usepackage{graphicx}
\usepackage{epstopdf}

\usepackage{wrapfig}
\usepackage{amsmath}

\title{Cascade Network with Guided Loss and Hybrid Attention for Two-view Geometry}

\author{
	Zhi Chen \and Fan Yang \and Wenbing Tao \thanks{Corresponding author}  \\
	\and National Key Laboratory of Science and Technology on Multi-spectral Information Processing\\
	\and School of Artifical Intelligence and Automation\\
	\and Huazhong University of Science and Technology, China \\
	\texttt{hust\_zhichen,hust\_fanyang,wenbingtao@hust.edu.cn} \\
}

\begin{document}

\maketitle

\begin{abstract}
 In this paper, we are committed to designing a high-performance network for two-view geometry.
 We first propose a Guided Loss and theoretically establish the direct negative correlation between the loss and Fn-measure by dynamically adjusting the weights of positive and negative classes during training, so that the network is always trained towards the direction of increasing Fn-measure. By this way, the network can maintain the advantage of the cross-entropy loss while maximizing the Fn-measure.
 We then propose a hybrid attention block to extract feature, which integrates the bayesian attentive context normalization (BACN) and  channel-wise attention (CA). BACN can mine the prior information to better exploit global context and CA can capture complex channel context to enhance the channel awareness of the network. 
 Finally, based on our Guided Loss and hybrid attention block, a cascade network \footnote{Our code will be available in Github later.} is designed to gradually optimize the result for more superior performance.
 Experiments have shown that our network achieves the state-of-the-art performance on benchmark datasets. 
\end{abstract}

\section{Introduction}
\label{introduction_section}
\begin{wrapfigure}{r}{7.5cm}
	\includegraphics[width=0.5\columnwidth]{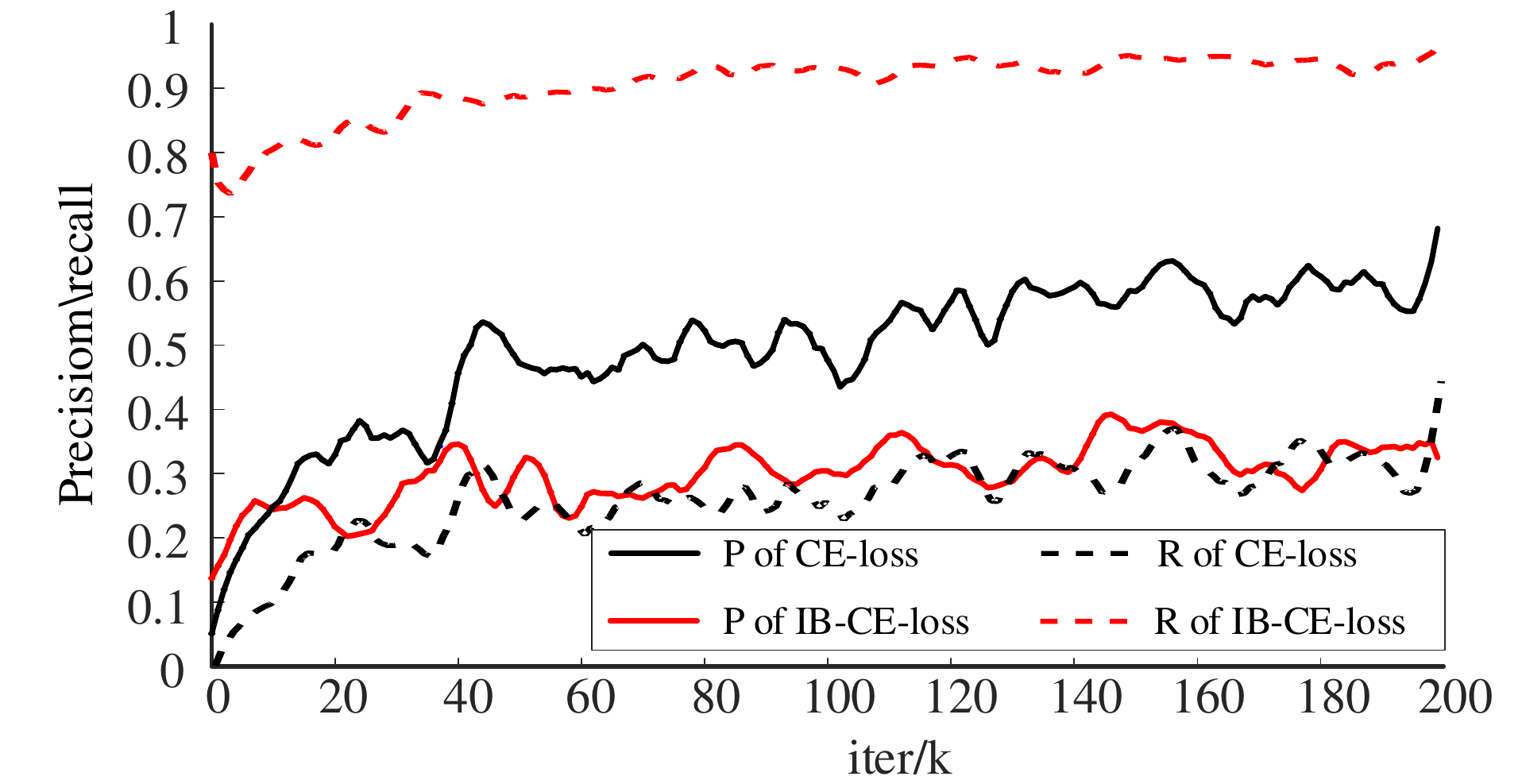}
	\caption{The training curves of different classification loss functions with same network. \textbf{CE-loss}: cross entropy loss. \textbf{IB-CE-loss}: instance balance cross entropy loss.}\label{introduction}
\end{wrapfigure}

Establishing stable correspondences and estimating two-view geometry between overlapping image pairs are the fundamental components of many tasks in computer vision, such as Structure from Motion (SfM) \cite{schonberger2016structure,snavely2006photo,agarwal2011building}, simultaneous localization and mapping (SLAM) \cite{benhimane2004real} and so on. Recently, some methods \cite{moo2018learning} adopt deep learning for two-view geometry. The network takes a putative correspondence set between an image pair as input and divides the putative correspondences into inliers (positive class) and outliers (negative class) to estimate essential matrix ($E$ matrix). Normally, the number of outliers in the putative correspondence set is much larger than inliers, which usually results in a class imbalance problem. In this case, it often happens that the loss function is too biased towards the positive or negative class because of the abnormal sample distribution. Thus, it is crucial to build a suitable loss function for two-view geometry task.

In order to show the relationship between loss functions and classification results, we train the same network (CN-Net \cite{moo2018learning} is used) with two commonly used loss functions and record the training curves of precision and recall as Fig. \ref{introduction} . When adopting cross entropy loss (CE-Loss), which considers each sample equally, the network is too biased towards negative class because the number of negative samples is larger than positive samples. Thus, the result of CE-loss maintains high precision and low recall. Instance balance cross entropy loss (IB-CE-Loss) \cite{deng2018pixellink} calculates the average loss of positive and negative classes respectively so that the proportion of positive and negative class in final loss will not be related to the number of samples. However, due to the number of positive samples is small, the proportion of each positive sample in the total loss is much larger than that of the negative sample. Thus, the cost of misclassifying a positive sample is so high that the result of IB-CE-loss is with high recall and low precision. In fact, the unbalanced precision and recall will lead to weak performance on Fn-measure \cite{van1974foundation}, which is the commonly used evaluation criterion of binary classification.

An immediate way of maximizing Fn-measure is directly using Fn-measure as objective function. In fact, Zhao et. al has already proposed to make Fn-measure differentiable and use it as loss function for salient object detection task \cite{zhao2019optimizing} . However, it may not be a good choice to directly use Fn-measure as objective function in the two-view geometry task, because CE-loss has been successfully applied to this topic \cite{moo2018learning,zhang2019learning,zhao2019nm} and proved to favor convergence of the network \cite{ng2017machine}. When replacing the IB-CE-loss of CN-Net \cite{moo2018learning} with Fn-measure, which has been verified in the subsequent experiments, the network does get a performance degradation. The degradation may  be caused by the following two reasons: 1) When using Fn-measure as objective function, some relaxation is necessary to make it differentiable. The relaxation may cause the network to deviate from the optimization goal. 2) If directly using Fn-measure as the objective function, the network cannot make use of all the samples, because the $TN$ (true negative) samples are not related with the computation of Fn-measure.

In order to retain the advantage of cross entropy loss while maximizing Fn-measure, we propose a new Guided Loss which keeps the form of the cross entropy loss and use the Fn-measure as a guidance to adjust the optimization goals dynamically. We theoretically prove that a perfect negative correlation can be established between the cross entropy loss and Fn-measure by dynamically adjusting the weights of positive and negative classes. Specifically, the perfect negative correlation is that the change in loss is completely opposite to the change in Fn-measure. Thus, with the decrease of the loss, the Fn-measure of the network will increase, so that the network is always trained towards the direction of increasing Fn-measure. By this way, the network maintains the advantage of the cross-entropy loss while maximizing the Fn-measure. It is worth mentioning when establishing the relationship between Fn-measure and loss, no relaxation is required.

Besides loss function, another challenge is how to better encode global context in the network. Unlike 3D point clouds, not each correspondence contributes to the global context. In contrast, outliers are noises to the global context. This issue is previously exploited by introducing spatial attention in the network \cite{Ploetz:2018:NNN,sun2019attentive}. However, the features of the shallow network are less recognizable, so it is hard to learn an appropriate weight vector when using self-attention operation in the shallow layers of the network. To address this issue, we propose a bayesian attentive context normalization (BACN) to mine prior information for better reducing the noise of outliers to global context. Besides, to capture more complex channel-wise context, we generalize the channel-wise attention (CA) \cite{hu2018squeeze} operation and reshape it as a point-wise form through group convolution \cite{cohen2016group}. The BACN and CA are further combined as a hybrid attention block for feature extraction.

Since the proposed Guided Loss can change the network's  bias toward precision and recall by using different Fn-measures (set $n$ as different value) as guidance, we can build a cascade network by means of the Guided Loss. Specifically, we first train the network through a Fn-measure with big $n$ as the guidance to obtain a coarse result with high recall. So the network keeps as many inliers as possible while filtering out some outliers. After that, Fn-measure with a smaller $n$ can be used as guidance to optimize the coarse result. As $n$ gets smaller, the network gradually leads to a result with higher precision. By gradually optimizing the result from coarse to fine, the network can achieve a better performance than that obtained by one fixed Fn-measure Guided Loss.

In a nutshell, our contribution is threefold: (i) We propose a novel Guided Loss for two-view geometry network. It can establish a direct connection between loss and Fn-measure, so the network can better optimize Fn-measure. (ii) We design a hybrid attention block to better extract global context. It combines a bayesian attentive context normalization and a channel-wise attention to capture the low-level prior information and channel-wise awareness. 
(iii) Based on the Guided Loss and hybrid attention block, we design a cascade network for two-view geometry estimation. Experiments  show that our network achieves state-of-the-art performance on benchmark datasets.

\section{Related Works}

\textbf{Model fitting methods} usually determine inliers by judging whether the raw matches satisfy the fitted epipolar geometric model. The classic RANSAC \cite{fischler1981random} adopts a hypothesize-and-verify pipeline, so do its variants, such as PROSAC \cite{chum2005matching}. Besides, many modifications of RANSAC have been proposed. Some methods \cite{chum2005matching,fragoso2013evsac,brahmachari2009blogs,goshen2008balanced} mine prior information to accelerate convergence. Some other methods \cite{chum2003locally,barath2018graph} augment the RANSAC by performing a local optimization step on the so-far-the-best model.



\textbf{Learning Based Methods.}
Since deep learning has been successfully applied for dealing with unordered data \cite{qi2017pointnet,qi2017pointnet++}, learning based methods attract great interest in two-view geometry estimation. 
CN-Net \cite{moo2018learning} reformulates the mismatch removal task as a binary classification problem. It utilizes a simple Context Normalization (CN) operation to extract global context. Based on CN, some network variants are proposed. NM-Net \cite{zhao2019nm} employs a simple graph architecture with an affine compatibility-specific neighbor mining approach to mine local context. $ \rm N^{3}$-Net \cite{Ploetz:2018:NNN} presents a continuous deterministic relaxtaion of KNN selection and a $\rm N^{3}$ block to mine non-local context. OA-Net \cite{zhang2019learning} utilizes an Order-Aware network to build model relation between different nodes. ACN-Net \cite{sun2019attentive} introduces spatial attention to two-view geometry network. 
Our work is to mine prior information and channel-wise awareness to improve the performance of the network.


\textbf{Attention Mechanism} focuses on perceiving salient areas similar to human visual systems \cite{vaswani2017attention}. Non-local neural network \cite{wang2018non} adopts non-local operation to introduce attention mechanism in feature map. SE-Net \cite{hu2018squeeze} introduces channel-wise attention mechanism through a Squeeze-and-Excitation block. In order to explore second-order statistics, SAN-Net \cite{dai2019second} utilizes second-order channel attention (SOCA) operations in their network. In addition to the two dimensional convolution, Wang et. al propose a graph attention convolution (GAC) \cite{wang2019graph} for dealing with point cloud data. 

\section{Method}
\subsection{Problem Formulation}
The input of our network is the coordinates of a set of putative correspondences, as follows: 
\begin{equation}\label{input}
C=[c_{1};c_{2};...,c_{N}] \in \mathbb{R}^{N \times 4}, c_{i}=(x_{1}^{i}, y_{1}^{i}, x_{2}^{i}, y_{2}^{i}),
\end{equation}
where $N$ is the number of putative correspondences. $(x_{1}^{i}, y_{1}^{i})$ and $(x_{2}^{i}, y_{2}^{i})$ are the coordinates of the two feature points of $i$-th correspondence. The coordinate of each feature point are normalized by camera intrinsics \cite{moo2018learning}. The network outputs the correspondence classification and $E$ matrix regression results in an end-to-end way, as follows:
\begin{equation}\label{network}
L = \Phi(C), L \in {\mathbb{R}}^{N \times 1}; \hat{E} = g(w, C), w=tanh({\rm ReLU}(L)), 
\end{equation}
where $\Phi(\cdot)$ is the network with trained parameters. $L$ is the logit values predicted by the network. $g(\cdot, \cdot)$ is the weighted eight-point algorithm \cite{moo2018learning} to estimate $E$ matrix, and $\hat{E}$ is the estimated $E$ matrix. 


\subsection{Guided Loss}
\label{headings}
The correspondence classification in our network is a binary classification task. In general, the result of binary classification is evaluated by the Fn-measure, as follows:
\begin{equation}\label{Fn}
Fn = {(1 + n^{2}) \cdot P \cdot R}/(n^{2} \cdot P + R).
\end{equation}
When $n$ > 1, the Fn-measure is biased in favour of recall and otherwise in favour of precision.
When adopting cross entropy loss as objective function, the loss will gradually decrease under the successive optimization. However, there is no guarantee that a drop in the loss will result in an increase of Fn-measure. Therefore, the network may not be trained towards the direction of optimizing Fn-measure. Based on this observation, we propose a hypothesis, that is, whether the relationship between the cross entropy loss and Fn-measure can be established, so that the decrease of loss will lead to the increase of Fn-measure.
This relationship can be expressed in the form of differential as follows:
\begin{equation}\label{dldf}
dloss \cdot dFn \leq  0.
\end{equation}
Specifically, we use the form of IB-CE-loss as follows:
\begin{equation}\label{rewritten_loss}
loss=-(\lambda \frac{1}{N_{pos}} \sum_{i=1}^{N_{pos}}\log(y_{i}) + \mu  \frac{1}{N_{neg}} \sum_{j=1}^{N_{neg}}\log(1-y_{j})), \\
s.t. \quad \ \lambda + \mu = 1, N_{pos} + N_{neg} = N 
\end{equation}
where $N_{pos}$ and $N_{neg}$ are the number of positive and negative samples. $\lambda$ and $\mu$ are the weights of positive and negative classes. 
Meanwhile, after forward propagation of the network, all the samples are divided into four categories, including $FP$ (false positive), $FN$ (false negative), $TP$ (true positive) and $TN$ (true negative). Suppose the number of $FN$ and $FP$ samples are $X$, $Y$ respectively, then the number of $TP$ and $TN$ can be computed as follows:
\begin{equation}\label{TPTN}
N_{TP} = {N_{pos} -X}, N_{TN} = {N_{neg} -Y},
\end{equation}
and the precision ($P$) and recall ($R$) in Fn-measure ($P$, $R$ in Eq. \ref{Fn}) can be computed as follows:
\begin{equation}\label{PR}
P = ({N_{pos} -X}) / ({N_{pos} - X + Y}),R = ({N_{pos} -X}) / {N_{pos}},
\end{equation}
Thus, Fn-measure is the dependent variable of X and Y according to Eq. \ref{Fn} and \ref{PR} . We express the functional relationship between Fn-measure ($Fn$) and $X, Y$ as follows:
\begin{equation}\label{F_X_Y}
Fn = {\rm {F}}(X, Y).
\end{equation}
In order to derive the relationship between Fn-measure and the loss, we also expect to express the loss as the dependent variables of X and Y. In the forward propagation of the network, we can calculate the average loss terms of $TP$, $TN$, $FP$ and $FN$ samples respectively, denoted as $l_{TP}, l_{TN}, l_{FP}, l_{FN}$. Then the loss in Eq. \ref{rewritten_loss} can be equivalently calculated as follows:
\begin{equation}\label{simplify_loss}
\begin{aligned}
loss = \lambda/N_{pos} \cdot \{X \cdot l_{FN} +
(N_{pos} - X) \cdot l_{TP}\} 
+ \mu/N_{neg} \cdot \{Y \cdot l_{FP} + (N_{neg} - Y) \cdot l_{TN}\}
\end{aligned}  
\end{equation}
We denote $dloss$ and $dFn$ as derivatives of X and Y as follows:
\begin{equation}\label{detail}
\begin{aligned}
dloss = \partial l_{X} dX + \partial l_{Y} dY, dFn = \partial F_{X} dX + \partial F_{Y} dY,
\end{aligned}
\end{equation}
where $\partial l_{X}$ and $\partial l_{Y}$ are the partial derivatives of loss with respect to $X$ and $Y$, and $\partial F_{X}$ and $\partial F_{Y}$ are the partial derivatives of Fn-measure with respect to $X$ and $Y$.
Then, we can draw a sufficient condition of Eq. \ref{dldf} as follows: 
\begin{equation}\label{st}
\partial F_{X}/{\partial F_{Y}} = \partial l_{X}/{\partial l_{Y}}.
\end{equation}

\textbf{Proof.} 
Since both the $TP$ and $FN$ samples belong to positive class (ground truth is positive), the loss term of each sample are computed by $-log(y_{i})$ in Eq. \ref{rewritten_loss} , where $y_{i}$ ($0 \leq y_i \leq 1$)  is the logit value. In fact, if the logit value of a positive sample (ground truth is positive) is greater than 0.5, then it is a $TP$ sample. And if the logit value is smaller than 0.5, it is a $FN$ sample. Obviously, since $-log(y_{i})$ is a monotone decreasing function, then the loss term of each $FN$ sample is greater than $TP$ sample. Thus, the average loss of $FN$ samples is greater than that of $TP$ samples, i. e.,  
\begin{equation}\label{constraint_loss_tmp1}
l_{FN} > l_{TP}.
\end{equation}
Similarly, the the average loss of $FP$ samples is greater than that of $TN$ samples, i. e., 
\begin{equation}\label{constraint_loss_tmp2}
l_{FP} > l_{TN}.
\end{equation}
Then, we can compute the partial derivatives of loss with respect to $X$ and $Y$ from Eq. \ref{simplify_loss}, as follows:
\begin{equation}\label{p_l_x}
\partial l_{X} = \lambda/N_{pos} \cdot (l_{FN} - l_{TP}), \partial l_{Y} = \mu/N_{neg}\cdot (l_{FP} - l_{TN})
\end{equation}
According to the constraints of Eq. \ref{constraint_loss_tmp1} and \ref{constraint_loss_tmp2}, we can obtain the following constraints:
\begin{equation}\label{constraint_loss}
\partial l_{X} > 0 ,\partial l_{Y} > 0.
\end{equation}
We then perform the same operations on $Fn$ to obtain the constraints of $Fn$. Specifically, we can compute the $\partial F_{X}$ and $\partial F_{Y}$ through compound derivation formula as follows:
\begin{equation}\label{f_x}
\partial F_{X} = \partial F_{P} \cdot \partial P_{X} + \partial F_{R} \cdot \partial R_{X},
\partial F_{Y} = \partial F_{P} \cdot \partial P_{Y} + \partial F_{R} \cdot \partial R_{Y}.
\end{equation}
The $\partial F_{P}$, $\partial F_{R}$, $\partial P_{X}$, $\partial P_{Y}$ all can be computed from Eq. \ref{Fn} and \ref{PR}, as follows:
\begin{equation}\label{f_p_x}
\begin{aligned}
\partial F_{P} = (1 + n^{2}) \cdot R^{2} / {(n^{2}P + R)^{2}} \geq 0, & \partial F_{R} = n^{2}(1 + n^{2}) \cdot P^{2} / {(n^{2}P + R)^{2}} \geq 0, \\
\partial P_{X} = -Y / {(N_{pos}-X+Y)^{2}} \leq 0, \partial P_{Y} & = -(N_{pos}-X) / {(N_{pos}-X+Y)^{2}} \leq 0, \\
\partial R_{X} = -1 / {N_{pos}} & \leq 0, \partial R_{Y} = 0,
\end{aligned}
\end{equation}
We can easily obtain the following constraints of $\partial F_{X}$ and $\partial F_{Y}$ from Eq. \ref{f_x}, \ref{f_p_x}:
\begin{equation}\label{constraint_fn}
\partial F_{X}\leq 0 ,\partial F_{Y} \leq 0.
\end{equation}
Meanwhile, according to Eq. \ref{detail}, we can get following equation:
\begin{equation}\label{dldf_expand}
\begin{aligned}
dloss \cdot dFn = & (\partial l_{X} dX + \partial l_{Y} dY) \cdot (\partial F_{X} dX + \partial F_{Y} dY) \\
= & \partial l_{X} \cdot \partial F_{X} \cdot (dX)^{2} + \partial l_{Y} \cdot \partial F_{Y} \cdot (dY)^{2}  + 
(\partial l_{X} \cdot \partial F_{Y} + \partial l_{Y} \cdot \partial F_{X}) \cdot dX \cdot dY
\end{aligned}
\end{equation}
According to the constraints of Eq. \ref{constraint_loss} and \ref{constraint_fn}, we can further expand Eq. \ref{dldf_expand} as follows:
\begin{equation}\label{dldf_expand2}
\begin{aligned}
dloss \cdot dFn = & 
\partial l_{X} \cdot \partial F_{X} \cdot (dX)^{2} + \partial l_{Y} \cdot \partial F_{Y} \cdot (dY)^{2} + 
(\partial l_{X} \cdot \partial F_{Y} + \partial l_{Y} \cdot \partial F_{X}) \cdot dX \cdot dY \\ 
= & \{\partial l_{X} \cdot \partial F_{X} \cdot (dX)^{2} + \partial l_{Y} \cdot \partial F_{Y} \cdot (dY)^{2} - 2 \sqrt{\partial l_{X} \cdot \partial F_{X} \cdot \partial l_{Y} \cdot \partial F_{Y}} \cdot dX \cdot dY\} \\
& + 
\{(\partial l_{X} \cdot \partial F_{Y} + \partial l_{Y} \cdot \partial F_{X}) \cdot dX \cdot dY + 2 \sqrt{\partial l_{X} \cdot \partial F_{X} \cdot \partial l_{Y} \cdot \partial F_{Y}} \cdot dX \cdot dY\
\} \\ 
= & - (\sqrt{-\partial l_{X} \cdot \partial F_{X}} \cdot dX + \sqrt{-\partial l_{Y} \cdot \partial F_{Y}}\cdot dY)^{2} - (\sqrt{-\partial l_{X} \cdot \partial F_{Y}} 
- \\ 
& \sqrt{-\partial l_{Y} \cdot \partial F_{X}})^{2} \cdot dX \cdot dY
\end{aligned}
\end{equation}
If Eq. \ref{dldf} holds, then, 
\begin{equation}\label{constraint4}
\begin{aligned}
\sqrt{-\partial l_{X} \cdot \partial F_{Y}} - \sqrt{-\partial l_{Y} \cdot \partial F_{X}} = 0,
\end{aligned}
\end{equation}
then, 
\begin{equation}\label{constraint5}
\begin{aligned}
dloss &\cdot dFn = - (\sqrt{-\partial l_{X} \cdot \partial F_{X}} \cdot dX + \sqrt{-\partial l_{Y} \cdot \partial F_{Y}}\cdot dY)^{2} \leq 0
\end{aligned}
\end{equation}

Thus, we can proof that Eq. \ref{st} is a sufficient condition of Eq. \ref{dldf} .

\textbf{Algorithm.} 
The main idea of the Guided Loss Algorithm is to make Eq. \ref{st} hold during training, so that a relationship of Eq. \ref{dldf} between the loss and Fn-measure can always be established. Specifically, the $\partial l_{X}$ and $\partial l_{Y}$ can be computed according to Eq. \ref{simplify_loss} as follows:
\begin{equation}\label{p_l_x}
\partial l_{X} = \lambda/N_{pos} \cdot (l_{FN} - l_{TP}), \partial l_{Y} = \mu/N_{neg}\cdot (l_{FP} - l_{TN}).
\end{equation}
Meanwhile, $\partial F_{X}$ and $\partial F_{Y}$ can also be calculated by means of numerical derivatives (step 4 in Algorithm \ref{alg:G_Loss} ) in the training process. Obviously, to hold Eq. \ref{st} , the weights $\lambda$ and $\mu$ should be dynamically changed during training. This also reveals the problem of IB-CE-Loss using a fixed $\lambda$ and $\mu$ during training. In order to establish a relationship between loss and Fn-measure as Eq. \ref{dldf} , we designed a weight adjustment algorithm by making Eq.\ref{st} hold , as follows:
\begin{algorithm} [h]
	\caption{Guided Loss} 
	\label{alg:G_Loss} 
	{\bf Input:} The classification result after forward propagation\\
	{\bf Output:} Proportion of positive and negative loss:  $\lambda$ and $\mu$
	\begin{algorithmic}[1] 

		\FOR{$i = 0; i < Batch\_size; i ++$}
		\STATE Count the number of $N_{pos_{i}}$ and $N_{neg_{i}}$. Count the number $TP$, $TN$, $FP$, $TN$ samples as $N_{TP_{i}}$, $N_{FP_{i}}$, $N_{TN_{i}}$, $N_{FN_{i}}$, then $X_{i} = N_{FN_{i}}$, $Y_{i}=N_{FP_{i}}$.

		\STATE Compute the average loss of $TP$, $TN$, $FP$ and $FN$ samples as $l_{TP_{i}}$, $l_{TN_{i}}$, $l_{FP_{i}}$ and $l_{FN_{i}}$.
		
		\STATE Compute $\partial F_{X_{i}}$ and $\partial F_{Y_{i}}$: $\partial F_{X_{i}}$ = ${\rm {F}}(X_{i} + 1, Y_{i}) - {\rm {F}}(X_{i}, Y_{i})$, $\partial F_{Y_{i}}$ = ${\rm {F}}(X_{i}, Y_{i} + 1) - {\rm {F}}(X_{i}, Y_{i})$
		
		\STATE s.t. $\lambda_{i} + \mu_{i} = 1$ $\rightarrow$ compute $\lambda_{i}$ and $\mu_{i}$ according to Eq. \ref{st} , \ref{p_l_x} and step 2, 3 and 4
		\ENDFOR
		\STATE return $\lambda$, $\mu$
	\end{algorithmic} 
\end{algorithm}

Specifically, when a batch of training data is sent to the network, the first step is forward propagation. After the forward propagation, we can use algorithm \ref{alg:G_Loss} to get $\lambda$ and $\mu$ for making Eq. \ref{st} hold. Then we substitute $\lambda$ and $\mu$ into Eq. \ref{rewritten_loss} and perform back propagation.

\begin{figure*}[t]
	\centering
	\includegraphics[width=1\columnwidth]{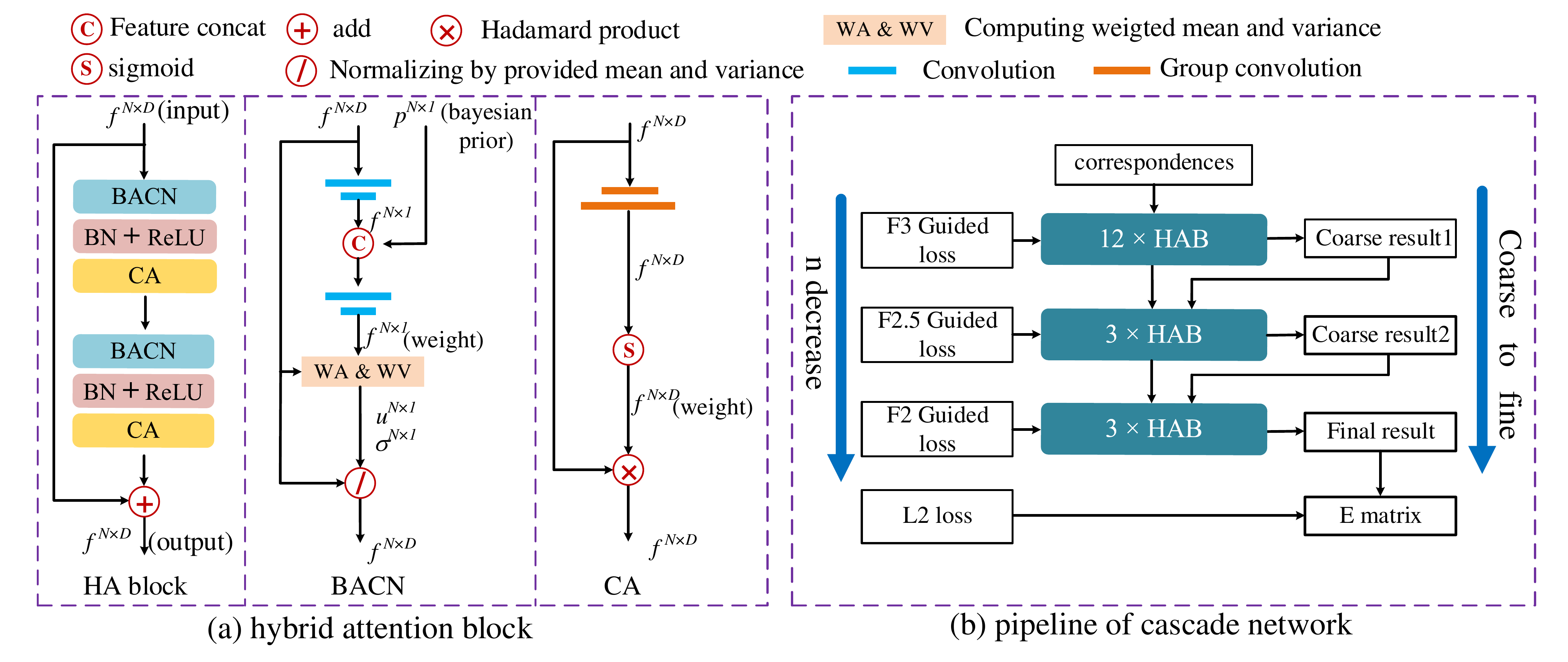} 
	\caption{\textbf{Network architecture}. (a) The hybrid attention block (HAB) is made up of bayesian attentive context normalization (BACN), batch normalization, ReLU and channel-wise attention (CA) in a Res-Net \cite{he2016deep} architecture. (b) The pipeline of the cascade network.} 
	\label{pipeline}
\end{figure*}

\subsection{Hybrid Attention Block.}
\textbf{Bayesian Attentive Context Normalization.} 
Specifically, suppose $f_{i}^{l} \in \mathbb{R}^{C}$ is the feature of correspondence $i$ in $l$-th layer, then the normalization operation in BACN can be expressed as follows:
\begin{equation}\label{BACN}
{\rm BACN}(f_{i}^{l}) = {(f_{i}^{l} - u^{l})} / {\sigma^{l}},
\end{equation}
where $u^{l}$ and $\sigma^{l}$ are the weighted mean and variance of all the features as shown in Fig. \ref{pipeline} (a) . We expect that the weights assigned to inliers in the weight vector for computing global context ($u^{l}$ and $\sigma^{l}$) is higher than that assigned to outliers, so that the impact of outliers on the global context are mitigated. To better learn the weight vector, we introduce the Lowe Ratio \cite{lowe2004distinctive} to generate a prior probability for each correspondence which can be used to assist the learning of weight vector.

Formally, given a pair of correspondence with Lowe's ratio $r_{i} \in \mathbb{R}^{1}$, we consider $r_{i}$ as a variable and the joint probability distribution function can be modeled as:
\begin{equation}
\begin{aligned}\label{Prior}
f_{r}(r_{i})=f_{in}(r_{i})\alpha+f_{out}(r_{i})(1-\alpha),
\end{aligned}
\end{equation}
where $f_{in}(r_{i})=f(r_{i}| r_{i}$ belongs to an inlier$)$, $f_{out}(r_{i})=f(r_{i}|r_{i}$ belongs to an outlier$)$, and $\alpha$ is the inlier ratio of the putative correspondence set of a specific image pair. 
Then, the prior probability $p_i(in)$ that the $i$-th correspondence belongs to inlier can be calculated as follows:
\begin{equation}
\begin{aligned}\label{Prior2}
p_i(in)=f_{in}(r_{i})\alpha / \{f_{in}(r_{i})\alpha + f_{out}(r_{i})(1-\alpha)\}.
\end{aligned}
\end{equation}
Before training, we obtain the empirically probability
density function of inlier ($f_{in}$) and outlier ($f_{out}$) respectively. Then for each image pair, we estimate the inlier ratio
$\alpha$ using a curve fitting method \cite{brahmachari2009blogs}. Thus we assign a prior probability of being an inlier to each correspondence by Eq. \ref{Prior2} . Finally, the prior probability participates in the calculation of weight vector as shown in Fig. \ref{pipeline} (a) .

\textbf{Channel-wise Attention.} The statistics on the channel have been shown to have a significant impact on the network \cite{hu2018squeeze}. In order to enhance the channel awareness of the network, we introduce channel-wise attention to the HA block. In order to capture complex channel context, we learn a channel weight vector for each correspondence instead of a weight vector that is shared by all the correspondences. When learning the weight vector, group convolution \cite{cohen2016group} is used to reduce network computation. Formally, Let $f_{i}^{l} \in \mathbb{R}^{C}$ be the feature of correspondence $i$ in $l$-th layer, then the CA can be expressed as follows: 
\begin{equation}
{\rm CA} (f_{i}^{l}) = f_{i}^{l} * w_{i}, i = 1,...N,   \\
\end{equation}
where $w_{i}$ is obtained by performing two group convolution operations \cite{cohen2016group} and a sigmoid function on the feature of $i$-th correspondence as shown in Fig. \ref{pipeline} (a).

\textbf{Hybrid Attention Block.} The BACN and CA are combined in Res-Net architecture as the feature extraction block, called hybrid attention block (HA Block) as Fig. \ref{pipeline} (a) . It is utilized as the basic feature extraction block in our network.
 
\subsection{Cascade Architecture.} 


Since the proposed Guided Loss can flexibly control the bias on precision and recall by using different Fn-measure as guidance, we can naturally build a cascade network  by Guided Loss to progressively refine the performance. 
Specifically, as shown in Fig. \ref{pipeline} (b) , we first use a 12-layer hybrid attention blocks as feature extraction module to extract the feature for each correspondence. Then a coarse result (coarse result1 in Fig. \ref{pipeline} (b)) can be obtained through these features by F3-measure Guided Loss. 
Then two refinement modules are followed to perform local optimization to refine the coarse result. 
Each refinement module is made up of a 3-layer HA Block. Different from feature extraction module, the global context in refinement module is extracted from the coarse result of the previous module instead of all of the correspondences. Besides, in order to gradually optimize the coarse result, the loss function will also progressively bias the precision. The coarse result2 is obtained by F2.5-measure Guided Loss, and the final result is obtained by F2-measure Guided Loss. Finally, the $E$ matrix is computed by performing weighted eight-point or RANSAC algorithm on the final result, and it is supervised by a $L2$-$loss$ .

\begin{table}
	\caption{Comparison with other baselines on $St\&Brown$ and $Colmap$ dataset. mAP (\%) (\textbf{with weighted eight-point algorithm/with RANSAC}) are reported.}
	\label{tab:overall_performance}
	\footnotesize
	\centering
	\begin{tabular}{lllllll}
		\toprule
		& \multicolumn{3}{c}{$St\&Brown$} & \multicolumn{3}{c}{$Colmap$}  \\                 
		\cmidrule(r){2-4} \cmidrule(r){5-7}
		& mAP $5^{\circ} $ & mAP $10^{\circ} $ & mAP $20^{\circ} $  & mAP $5^{\circ} $ & mAP $10^{\circ} $ & mAP $20^{\circ} $ \\
		\midrule
		RANSAC  & -/4.00 & -/6.82 & -/11.52 & -/2.28 & -/4.52 & -/5.68 \\
		CN-Net \cite{moo2018learning} & 15.12/33.11
		& 31.87/43.47 & 43.62/54.81 & 11.82/26.89 & 18.44/30.82 & 23.89/34.52  \\ 
		Point-Net++ \cite{qi2017pointnet++}  & 12.12/26.31 & 27.85/33.92 & 33.88/45.68 & 10.41/25.65 & 17.94/28.76 & 22.55/32.10 \\
		ACN-Net \cite{sun2019attentive} & 25.87/35.68 & 35.66/46.04 & 47.69/58.25 & 21.65/30.40 & 25.72/34.89 & 30.02/41.58 \\ 
		NM-Net \cite{zhao2019nm} & 17.70/34.09 & 32.80/42.92 & 43.74/54.62 & 20.96/31.72 & 23.08/33.42 & 29.18/39.62 \\ 
		$\rm N^{3}$-Net \cite{Ploetz:2018:NNN} & 14.52/32.65 & 30.27/42.16 & 40.84/52.69 & 10.90/25.68 & 16.74/29.77 & 23.11/34.09
		\\ 
		
		OA-Net \cite{zhang2019learning} & 30.53/37.80 & 39.84/49.87 & 50.01/61.91 & 26.82/34.57 & 29.99/37.09 & 34.98/45.54 
		\\ 
		
		Ours & \textbf{31.25/41.90} & \textbf{41.52/52.57} & \textbf{53.64/63.60} & \textbf{27.82/36.83} & \textbf{30.80/39.26} & \textbf{35.90/46.52} \\
		\bottomrule
	\end{tabular}
\end{table}

\textbf{Loss Function.} We formulate our training objective as a combination of two types of loss functions, classification loss and
regression loss. The whole objective function is as follows:
\begin{equation}
\begin{aligned}\label{wholeLoss}
loss=l_{cls}+\eta_{1} l_{cls1}+\eta_{1} l_{cls2}+\eta_{3} l_{reg}.
\end{aligned}
\end{equation} 
As shown in Fig. \ref{pipeline} , $l_{cls}$ is related with the final result, and $l_{cls1}$ and $l_{cls2}$ are related with the coarse result1 and coarse result2 respectively. For the regression loss $l_{reg}$, we use $L2$-$loss$ as follows:
\begin{equation}
\label{loss_reg}
l_{reg}=min\{\left \| \hat{E} \pm E \right \| \}, 
\end{equation} 
where $\hat{E}$ is the estimated $E$ matrix and $E$ is the ground truth $E$ matrix.


\section{Experiments}

\subsection{Experimental Setup}
\textbf{Parameter Settings.} 
The network is trained by Adam optimizer with a learning rate being $10^{-3}$ and batch size being 16. The iteration times are set to 200k. In Eq. \ref{wholeLoss} , the loss weight $\eta_{3}$ is 0 during the first 20k iteration and then 0.1 later. $\eta_{1}$ and $\eta_{2}$ are set to 0.1 during the whole training.

\textbf{Datasets.} 
Yi et. al \cite{moo2018learning} evaluate their approach on a dataset which contains outdoor and indoor scenes. They choose 5 scenes from the Structure from Motion (SfM) dataset as outdoor scene and 16 scenes from SUN3D dataset \cite{xiao2013sun3d} as indoor scene. They use the SfM pipeline \cite{wu2013towards} to generate  Ground Truth for outdoor scene and use KinectFusion \cite{newcombe2011kinectfusion} for indoor scene. Yi et al. kindly provided us with their datasets and their exact data splits. We use their dataset and their setup. Since the $St. Peter’s$ and the $Brown$ scenes are used as training data, we will abbreviate this dataset as the $St\&Brown$ dataset in this paper. Besides, zhao et. al \cite{zhao2019nm} also provide a outdoor dataset, which contains 16 outdoor scenes. They also kindly provide us with their datasets and their exact data splits. We call it as $Colmap$ dataset because some datas are from the Colmap dataset \cite{schonberger2016structure}. Note that training and test are performed on completely separate scenes so that the  network has to generalize to unknown environments.

\textbf{Evaluation Criteria.} 
To measure the performance of final result ($E$ matrix regression), we recover the rotation and translation vector from estimated $E$ matrix, and then use the angular error between the ground-truth and estimated value of both rotation and translation \cite{moo2018learning}. The mAP under $5^{\circ}$, $10^{\circ}$, $20^{\circ}$ are all reported as the metrics.

\begin{figure*}
	\centering
	\subfigure[weight $\lambda$ ]{\includegraphics[width=2.7in]{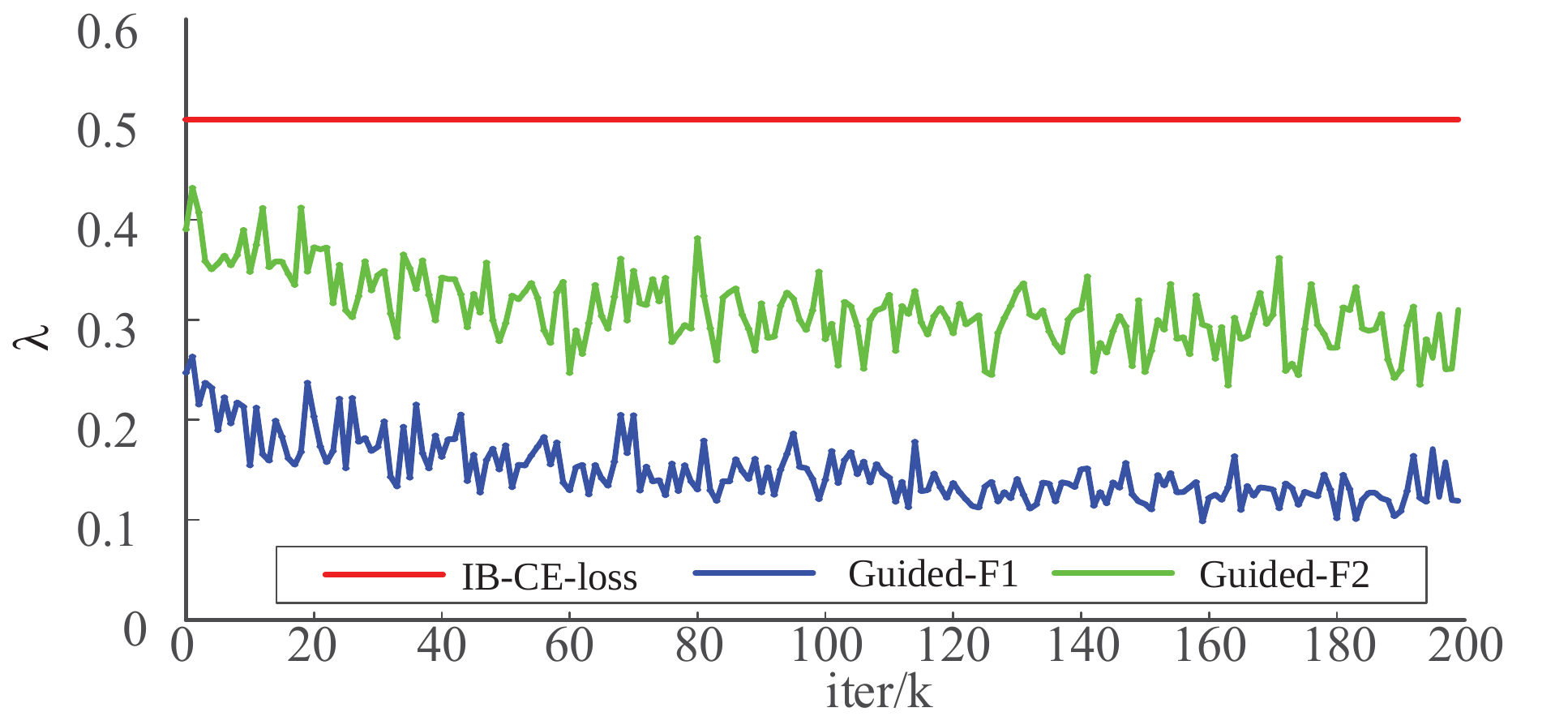}}
	\subfigure[Precision and recall]{\includegraphics[width=2.7in]{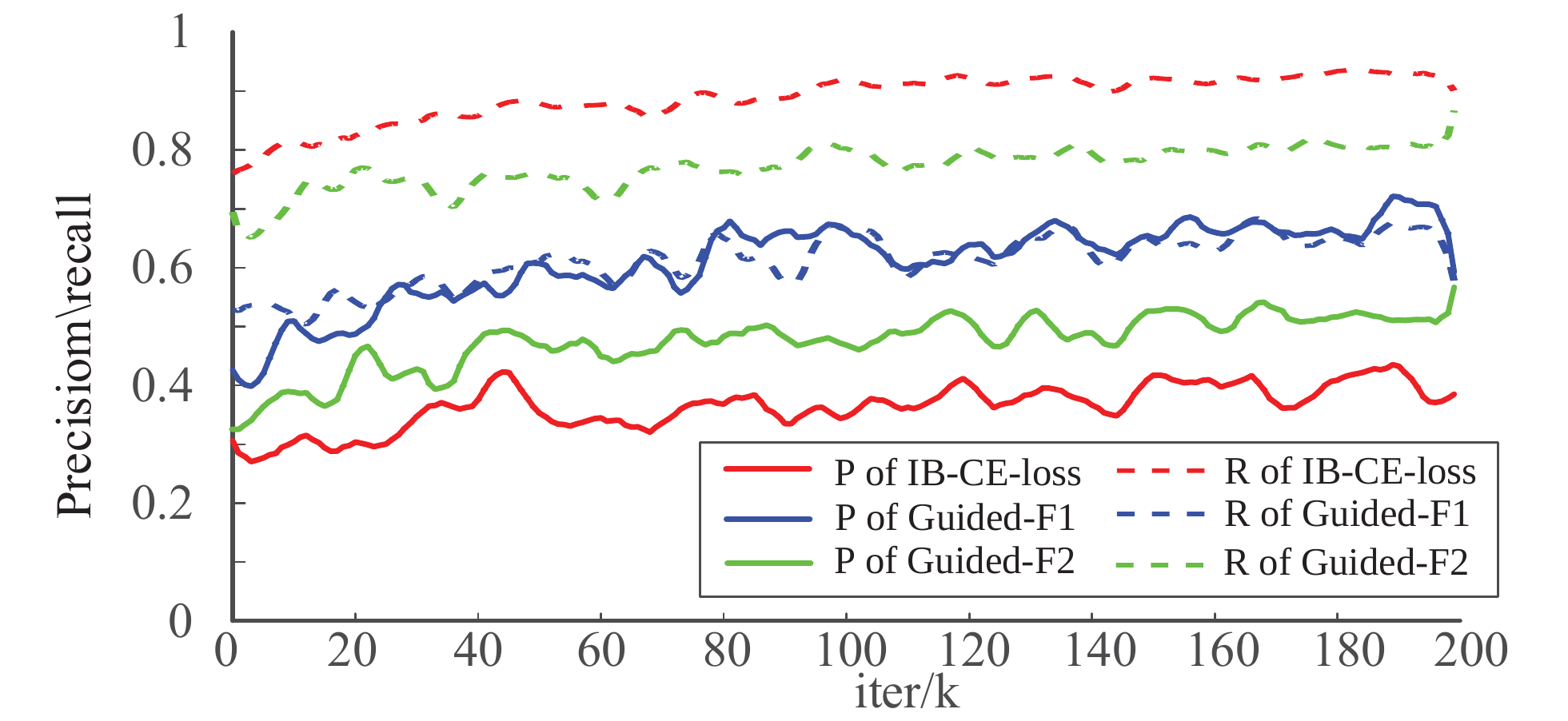}}
	\caption{\textbf{Training curves.} We train the baseline CN-Net \cite{moo2018learning} with different classification loss on $St\&Brown$ dataset. (a) The weight curve of positive class ($\lambda$ in Eq. \ref{rewritten_loss}). (b) The precision and recall curves on the  validation set. }
	\label{training_curve}
\end{figure*}

\begin{table*}[t]
	\centering%
	\caption{Ablation study on $St\&Brown$ datasets. mAP (\%) under $5^{\circ}$ and $10^{\circ}$ (using \textbf{weighted eight-point/RANSAC} as post-processing) are reported.  \textbf{BACN}: bayesian attention context normalization. \textbf{CA}: channel-wise attention. \textbf{ACN}: attentive context normalization \cite{sun2019attentive}. \textbf{G-Loss}: Guided Loss. \textbf{No cas}: 18-layer network without cascade architecture. \textbf{Cas:} 18-layer network with cascade architecture.} \label{tab:Ablation}%
	\footnotesize
	
	\begin{tabular}{llllllllll}
		\toprule
		\multicolumn{7}{c}{module} & \multicolumn{2}{c}{result}  \\                 
		\cmidrule(r){1-7} \cmidrule(r){8-9}
		CN-Net & BACN & CA & ACN &  G-Loss & No-cas & cas & mAP $5^{\circ} (\%)$ & mAP $10^{\circ} (\%)$\\ 
		\checkmark & & & & & &  & 15.12/33.11
		& 31.87/43.47 \\ 
		\checkmark & \checkmark &  & & & & & 26.32/36.14 & 36.41/46.88 \\ 
		
		\checkmark & & \checkmark & & & & & 26.91/36.85 & 37.02/47.48 \\ 
		\checkmark & \checkmark & \checkmark & & & & & 27.83/37.99 & 37.82/48.29 \\
		
		\checkmark & & & \checkmark & & & &  25.87/35.68 & 35.66/46.04 \\ 
		\checkmark & \checkmark & \checkmark & & \checkmark & & & 29.95/39.88 & 39.38/50.02 \\ 
		
		\checkmark & \checkmark & \checkmark & & \checkmark & \checkmark & & 30.32/40.18 & 40.25/50.51\\
		
		\checkmark & \checkmark & \checkmark & & \checkmark & & \checkmark &  {31.25/41.90} & 41.52/52.57 \\
		\bottomrule
	\end{tabular}

\end{table*}

\subsection{Comparison to Other Baselines}
In order to verify the performance of Guided Loss in the network, we record the training curves of the weight, precision and recall in Fig. \ref{training_curve} . Since the sum of the weight $\lambda$ of the positive class and the weight $\mu$ of the negative class  in loss function is always 1, we only record the curve of $\lambda$. As shown in Fig. \ref{training_curve} (a),  $\lambda$ in the Guided Loss is dynamically changed, while $\lambda$ in IB-CE-Loss is set to fixed value 0.5. As a result, the Guided Loss can achieve a balance between precision and recall, as shown in Fig. \ref{training_curve} (b). Meanwhile, when using F1-measure, which considers precision and recall equally, as the guidance, the gap between precision and recall is always small. And when using F2-measure, which is more bias towards recall, the recall is always higher than precision during training. It shows that the result of Guided Loss always accords with the guided Fn-measure, which verifies the effect of the guidance. 

\begin{figure*}[t]
	\centering
	\includegraphics[width=1\columnwidth]{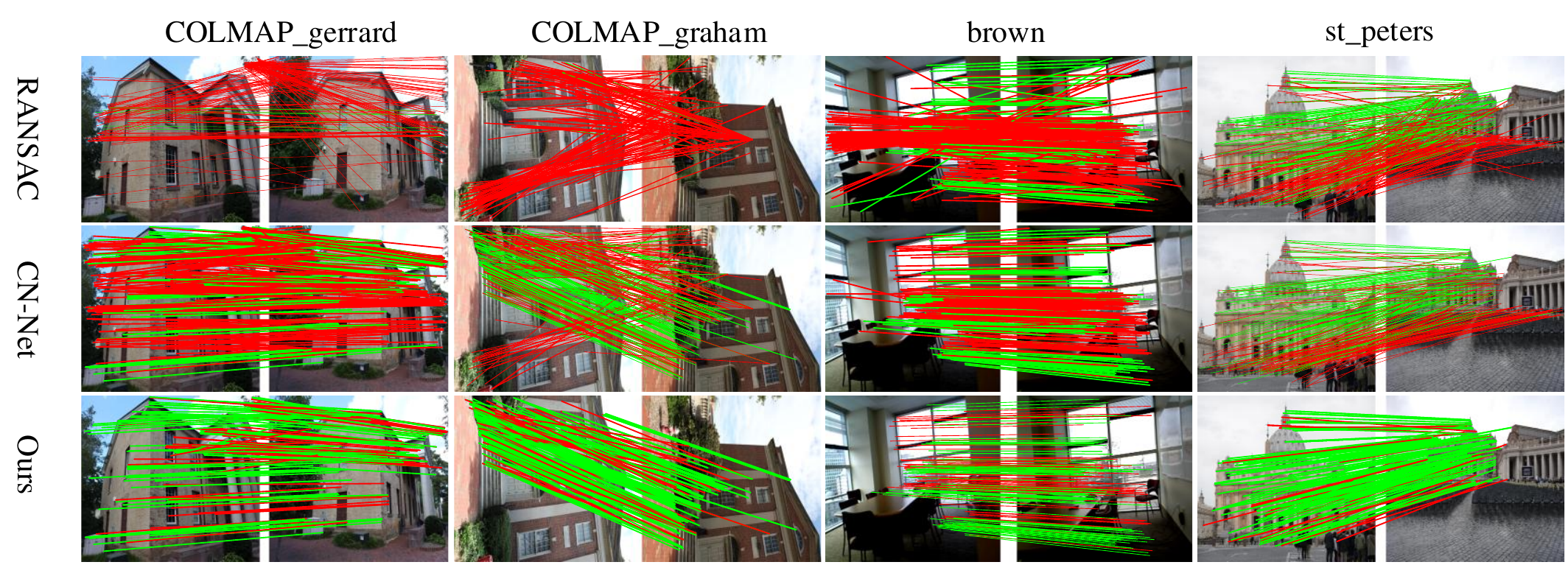} 
	\caption{Visual comparison of matching results using RANSAC, CN-Net and our method. Images are taken from $Colmap$ and $St\&Brown$ datasets. Correspondences are in green if they are inliers, and in red otherwise. \textbf{Best viewed in color.}} 
	\label{visualize}
\end{figure*}

Then we compare our network with other state-of-the-art
networks \cite{moo2018learning,qi2017pointnet++,sun2019attentive,zhao2019nm,Ploetz:2018:NNN,zhang2019learning} on both $St\&Brown$ and $Colmap$ datasets. All the networks are trained with the same setting. The weighted eight-point and RANSAC methods are utilized as post-processing, respectively, and the mAP under $5^{\circ}$, $10^{\circ}$ and $20^{\circ}$ are reported in Tab. \ref{tab:overall_performance}. Our approach significantly exceeds our baseline network (CN-Net) over 10\% on both of $St\&Brown$ and $Colmap$ dataset, and achieves a better performance than the other networks. It is worth noting that our network works better when using RANSAC as a post-processing method. This is because RANSAC allows a certain proportion of mismatches in the original match set. Thus, the more matches the network retains within the RANSAC's anti-noise range, the more accurate the $E$ matrix estimation will be. We choose F2-measure Guided loss as the objective function, so that the network can not only ensure an acceptable precision, but also make the recall as high as possible. A visualized matching results of RANSAC \cite{fischler1981random}, CN-Net \cite{moo2018learning} and our network in Fig. \ref{visualize}.

\begin{table}
	\caption{The results of three networks with both their original classification loss (without "+" in the table) and our F2-measure Guided Loss (with "+" in the table). mAP (\%) $5^{\circ}$, $10^{\circ}$ and $20^{\circ}$ (using \textbf{weighted eight-point/RANSAC} as post-processing) are reported.}
	\label{tab:overall_performance}
	\footnotesize
	\centering
	\begin{tabular}{lllllll}
		\toprule
		& \multicolumn{3}{c}{$St\&Brown$} & \multicolumn{3}{c}{$Colmap$}  \\                 
		\cmidrule(r){2-4} \cmidrule(r){5-7}
		& mAP $5^{\circ} $ & mAP $10^{\circ} $ & mAP $20^{\circ} $  & mAP $5^{\circ} $ & mAP $10^{\circ} $ & mAP $20^{\circ} $ \\
		\midrule
		CN-Net \cite{moo2018learning} &  15.12/33.11
		& 31.87/43.47 & 43.62/54.81 & 11.82/26.89 & 18.44/30.82 & 23.89/34.52  \\
		CN-Net + & 18.53/34.80
		& 33.41/45.19 & 45.11/56.10 & 12.54/27.74 & 19.62/31.91 & 25.11/36.01  \\ 
		ACN-Net \cite{sun2019attentive}   & 25.87/35.68 & 35.66/46.04 & 47.69/58.25 & 21.65/30.40 & 25.72/34.89 & 30.02/41.58 \\
		ACN-Net + & 27.32/37.13 & 36.98/47.65 & 49.09/60.18 & 22.78/31.76 & 26.85/36.21 & 31.56/42.09 \\ 
		NM-Net \cite{zhao2019nm} & 17.70/34.09 & 32.80/42.92 & 43.74/54.62 & 20.96/31.72 & 23.08/33.42 & 29.18/39.62 \\ 
		
		NM-Net + & 19.89/36.79 & 34.32/44.02 & 44.85/56.17 & 21.78/32.43 & 24.05/34.87 & 30.22/40.74 
		\\ 
		
		\bottomrule
	\end{tabular}
\end{table}

\subsection{Ablation studies}

\textbf{HA Block \textit{vs.} ACN Block\cite{sun2019attentive}.} To demonstrate the performance
of HA block, we replace the CN Block in the baseline CN-Net \cite{moo2018learning} with the HA block. Both the bayesian attentive context normalization (BACN) and channel-wise attention (CA) are tested specifically as Tab. \ref{tab:Ablation} . As a comparison, we also replace CN Block with ACN Block \cite{sun2019attentive} to train the network. Both of the BACN and CA achieve a better result than ACN, and HA block (BACN + CA) achieves an improvement of about 2\% over ACN in both mAP $5^{\circ}$ and $10^{\circ}$.

\textbf{Guided Loss.} 
We replace the IB-CE-Loss with our F2-measure Guided Loss. As shown in Tab. \ref{tab:Ablation} , the proposed Guided Loss (CN-Net + BACN + CA + G-Loss) achieves a better performance over the original loss of CN-Net (CN-Net + BACN + CA) about 2\%.

\textbf{Cascade \textit{vs.} No Cascade.} 
In order to show the performance of the proposed cascade architecture, we first deepen the layers of CN-Net from 12 to 18 and test the result as comparison.
Meanwhile, we also train the proposed cascaded network, which is also a 18-layer network.  As shown in Tab. \ref{tab:Ablation} , only increasing the number of network layers, the performance of the network is not significantly improved. The performance of cascade network with the same number of layers is significantly better than non-cascaded networks. It implies that using the Guided Loss in a coarse-to-fine cascade manner can significantly improve network performance.

\begin{wraptable}{r}{8.2cm}
	\centering%
	\caption{mAP (\%) $5^{\circ}$, $10^{\circ}$ and $20^{\circ}$ (using \textbf{weighted eight-point/RANSAC} as post-processing) on the $St\&Brown$ dataset of different classification loss functions. Fn-Loss ($n$ = 1, 2) is using Fn-measure as objective function, while Guided Fn ($n$ = 1, 2) is the proposed Guided Loss.} \label{tab:with_other_loss}%
	\footnotesize
	
	\begin{tabular}{llll}
		\toprule
		& mAP $5^{\circ} $ & mAP $10^{\circ} $ & mAP $20^{\circ} $ \\
		\midrule
		CE-Loss & 10.12/26.31 & 17.85/35.92 & 32.14/51.92 \\
		IB-CE-Loss \cite{deng2018pixellink} & 15.12/33.12 & 22.65/42.97 & 34.32/54.63 \\ 
		Focal Loss \cite{lin2017focal} & 11.32/27.65 & 18.94/37.43 & 32.04/52.10 \\
		F1-Loss \cite{zhao2019optimizing} & 9.82/26.90 & 16.16/38.27 & 26.83/53.32 \\ 
		F2-Loss \cite{zhao2019optimizing} & 8.64/28.72 & 14.07/39.28 & 23.84/51.25 \\ 
		Guided F1 & 15.90/33.42 & 23.98/43.57 & 35.69/55.32
		\\
		
		Guided F2 & \textbf{18.52/34.83} & \textbf{25.41/44.64} & \textbf{36.98/56.12}
		\\ 
		\bottomrule
	\end{tabular}
\end{wraptable}

%

\subsection{Guided Loss}	
\textbf{Guided Loss \textit{vs.}another loss.} In order to verify the effectiveness of our Guided Loss, we train the CN-Net \cite{moo2018learning} with the different loss functions \cite{deng2018pixellink,zhao2019optimizing,lin2017focal} and the mAP under $5^{\circ}$, $10^{\circ}$ and $20^{\circ}$  are reported in Tab. \ref{tab:with_other_loss} . As discussed in Section \ref{introduction_section}, when using Fn-measure as objective function, some relaxation has to be made and not all of the samples are utilized for back propagation. Therefore, Fn-Loss does not even perform as well as IB-CE-Loss. For the proposed Guided Loss, the network can achieve a better result than the other loss functions whether F1-measure or F2-measure is used as guidance.
This is because the Guided Loss can maintain the advantages of IB-CE-Loss while achieving a balance between precision and recall. Note that Focal-Loss is designed to mine hard samples, so it does not perform well in this case. 

\textbf{Guided Loss with other baseline networks.}  We further analyze our Guided Loss by replacing the classification loss functions of other models with Guided Loss. We first train three recent networks, including CN-Net \cite{moo2018learning}, ACN-Net \cite{sun2019attentive} and NM-Net \cite{zhao2019nm}, with their original classification loss. Then we replace their classification loss with our F2-measure Guided Loss. The results are reported in Tab. \ref{tab:overall_performance}. Each network with the supervision of our loss can increase the mAP by 1-3\% without modifying anything.

\section{Conclusion}
In this paper, a novel Guided Loss is proposed to build direct link between loss function and the evaluation criterion, i. e., Fn-measure, for training the network to better optimize Fn-measure. With different Fn-measure as guidance, one can easily adjust the compromise between precision and recall, enabling flexibility to deal with various
applications. Besides, a hybrid attention (HA) block, including a bayesian attentive context normalization and a channel-wise attention, is proposed for better extracting global context. The Guided Loss and HA Block are combined in a cascade network for two-view geometry tasks. Through extensive experiments, we demonstrate that our network can achieve the state-of-the-art performance on benchmark dataset.

\small
\bibliographystyle{IEEEtran}
\bibliography{IEEEabrv,egbib}

\end{document}